\documentclass[10pt,twocolumn,letterpaper]{article}

\usepackage{cvpr}              

\usepackage[accsupp]{axessibility}  
\usepackage{graphicx}
\usepackage{amsmath}
\usepackage{amssymb}
\usepackage{booktabs}
\usepackage{times}
\usepackage{epsfig}
\DeclareMathOperator*{\argmin}{arg\,min}
\DeclareSymbolFont{matha}{OML}{txmi}{m}{it}
\DeclareMathSymbol{\varv}{\mathord}{matha}{118}
\usepackage{array}
\usepackage{tabularx}
\usepackage{multirow, booktabs}
\usepackage{float}
\usepackage{bm}
\usepackage[dvipsnames,table,xcdraw]{xcolor}
\usepackage{resizegather}

\usepackage[pagebackref,breaklinks,colorlinks]{hyperref}

\usepackage[capitalize]{cleveref}
\crefname{section}{Sec.}{Secs.}
\Crefname{section}{Section}{Sections}
\Crefname{table}{Table}{Tables}
\crefname{table}{Tab.}{Tabs.}

\def\cvprPaperID{****} 
\def\confName{CVPR}
\def\confYear{2022}

\begin{document}

\title{Affine Medical Image Registration with Coarse-to-Fine Vision Transformer}

\author{Tony C. W. Mok, Albert C. S. Chung\\
Department of Computer Science and Engineering,\\
The Hong Kong University of Science and Technology\\
{\tt\small cwmokab@connect.ust.hk, achung@cse.ust.hk}
}

\maketitle

\begin{abstract}
Affine registration is indispensable in a comprehensive medical image registration pipeline. However, only a few studies focus on fast and robust affine registration algorithms. Most of these studies utilize convolutional neural networks (CNNs) to learn joint affine and non-parametric registration, while the standalone performance of the affine subnetwork is less explored. Moreover, existing CNN-based affine registration approaches focus either on the local misalignment or the global orientation and position of the input to predict the affine transformation matrix, which are sensitive to spatial initialization and exhibit limited generalizability apart from the training dataset. In this paper, we present a fast and robust learning-based algorithm, Coarse-to-Fine Vision Transformer (C2FViT), for 3D affine medical image registration. Our method naturally leverages the global connectivity and locality of the convolutional vision transformer and the multi-resolution strategy to learn the global affine registration. We evaluate our method on 3D brain atlas registration and template-matching normalization. Comprehensive results demonstrate that our method is superior to the existing CNNs-based affine registration methods in terms of registration accuracy, robustness and generalizability while preserving the runtime advantage of the learning-based methods. The source code is available at \url{https://github.com/cwmok/C2FViT}. 
\end{abstract}

\begin{figure}[t]
	\begin{center}
        \includegraphics[width=1.0\linewidth]{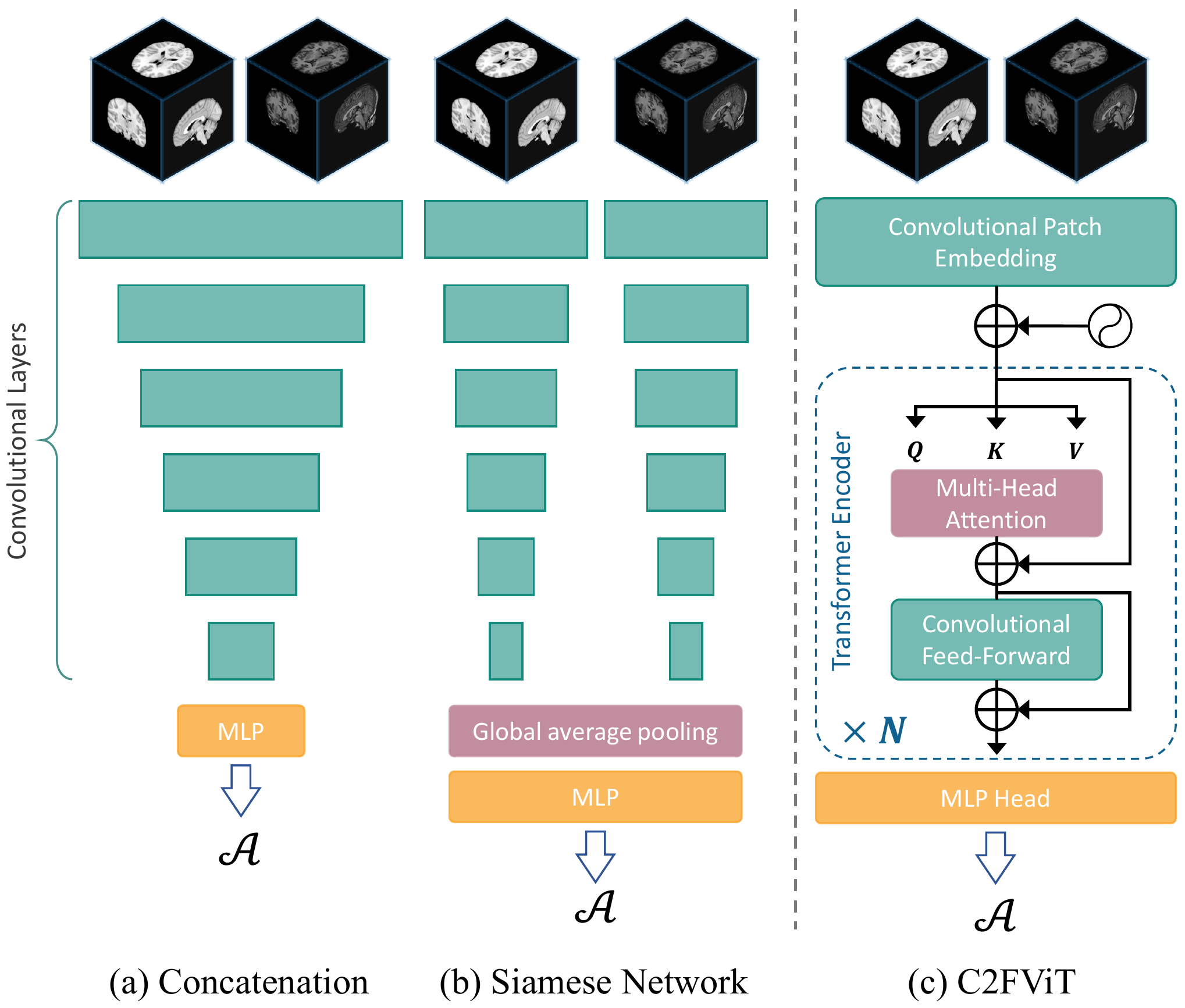}
	\end{center}
	\caption{Comparisons of different architectures for affine registration. The concatenation-based (VTN-Affine\cite{zhao2019unsupervised}) and Siamese network (ConvNet-Affine\cite{de2019deep}) approaches are based on convolutional neural networks, while our proposed C2FViT is based on vision transformers. For brevity, we illustrate 1-level C2FViT only. Local and global operations are in green and purple, respectively.}

	\label{fig:compare}
\end{figure}

\section{Introduction}
Rigid and affine registration is crucial in a variety of medical imaging studies and has been a topic of active research for decades. In a comprehensive image registration framework, the target image pair is often pre-aligned based on a rigid or affine transformation before using deformable (non-rigid) registration, eliminating the possible linear and large spatial misalignment between the target image pair. Solid structures such as bones can be aligned well with rigid and affine registration \cite{maintz1998survey,pluim2003mutual}. In conventional image registration approaches, inaccurate pre-alignment of the image pair may impair the registration accuracy or impede the convergence of the optimization algorithm, resulting in sub-optimal solutions \cite{zhou2014novel}. The success of recent learning-based deformable image registration approaches has largely been fueled \cite{de2019deep,balakrishnan2018unsupervised,dalca2018unsupervised,heinrich2019closing,hering2021cnn,mok2020fast,mok2020large,mok2021conditional,hoopes2021hypermorph} by accurate affine initialization using conventional image registration methods. While the conventional approaches excel in registration performance, the registration time is dependent on the degree of misalignment between the input images and can be time-consuming with high-resolution 3D image volumes. To facilitate real-time automated image registration, a few studies \cite{zhao2019unsupervised,shen2019networks,hu2018label,huang2021coarse} have been proposed to learn joint affine and non-parametric registration with convolutional neural networks (CNNs). However, the standalone performance of the affine subnetwork compared to the conventional affine registration algorithm is less explored.
Moreover, considering that affine transformation is global and generally targets the possible large displacement, we argue that CNNs are not the ideal architecture to encode the orientation and absolution position of the image scans in Cartesian space or affine parameters due to the inductive biases embedded into the architectural structure of CNNs. 

In this paper, we analyze and expose the generic inability and limited generlizability of CNN-based affine registration methods in cases with large initial misalignment and unseen image pairs apart from the training dataset. Motivated by the recent success of vision transformer models \cite{vaswani2017attention,dosovitskiy2020image,wang2021pyramid,d2021convit,wu2021cvt}, we depart from the existing CNN-based approaches and propose a coarse-to-fine vision transformer (C2FViT) dedicated to 3D medical affine registration. To the best of our knowledge, this is the first learning-based affine registration approach that considers the non-local dependencies between input images when learning the global affine registration for 3D medical image registration. 

The main contributions of this work are as follows:
\begin{itemize}
	\item we quantitatively investigate and analyze the registration performance, robustness and generalizability of existing learning-based affine registration methods and conventional affine registration methods in 3D brain registration;
	\item we present a novel learning-based affine registration algorithm, namely C2FViT, which leverages convolutional vision transformers with the multi-resolution strategy. C2FViT outperforms the recent CNN-based affine registration approaches while demonstrating superior robustness and generalizability across datasets; 
	\item the proposed learning paradigm and objective functions can be adapted to a variety of parametric registration approaches with minimum effort.
\end{itemize}

We evaluate our method on two tasks: template-matching normalization to MNI152 space \cite{grabner2006symmetric,evans2012brain,fischl2012freesurfer} and 3D brain atlas registration in native space. Results demonstrate that our method not only achieves superior registration performance over existing CNN-based methods, but the trained model also generalizes well to an unseen dataset beyond the training dataset, reaching the registration performance of conventional affine registration methods.

\section{Related Work}
\subsection{Learning-based Affine Registration Methods}
Conventional approaches often formulate the affine registration problem to an iterative optimization problem, which optimizes the affine parameters directly using adaptive gradient descent \cite{klein2009elastix,avants2009advanced} or convex optimization \cite{heinrich2015multi}. While conventional approaches excel in registration accuracy, the registration time is subject to the complexity and resolution of the input image pairs. Recently, many learning-based approaches have been proposed for fast affine registration. These approaches significantly accelerate the registration time by formulating the affine registration problem as a learning problem using CNNs and circumventing the costly iterative optimization in conventional approaches. Existing CNN-based affine registration approaches can be divided into two categories: concatenation-based \cite{zhao2019unsupervised,hu2018label,huang2021coarse,miao2016cnn} and Siamese network approaches \cite{de2019deep,chen2021learning,shao2021weakly} as shown in figure \ref{fig:compare}.

Zhao et al. \cite{zhao2019unsupervised} propose a concatenation-based affine subnetwork that concatenates the fixed and moving images as input, and exploits single-stream CNNs to extract the features based on the local misalignment of the input. Considering affine registration is global, their method is not capable of input with large initial misalignment as the affine subnetwork lacks global connectivity and only focuses on the overlapping region between two image spaces. In contrast to the concatenation-based method, de Vos et al. \cite{de2019deep} propose an unsupervised affine registration method using the Siamese CNN architecture for fixed and moving images. A global average pooling \cite{lin2013network} is applied to the end of each pipeline in order to extract one feature per feature map, forcing the networks to encode orientations and affine transformations globally. Although their network focuses on the global high-level geometrical features of separated input, their method completely ignores the local features of the initial misalignment between the input image pair. Moreover, a recent study \cite{liu2018intriguing} demonstrates that a pure CNN encoder fails spectacularly in a seemingly trivial coordinate transform problem, implying that a pure CNN encoder may not be an ideal architecture to encode the orientations and absolution positions of the image scans in Cartesian space or to affine parameters. Shen et al. \cite{shen2019networks} also report that CNN-based affine registration methods do not perform well in practice, even for deep CNNs with large receptive fields.

It is worth noting that most of the existing CNN-based affine registration methods \cite{de2019deep,zhao2019unsupervised,hu2018label,huang2021coarse,shao2021weakly,chen2021learning} jointly evaluate the affine and deformable registration performance or completely ignore the standalone performance of the affine subnetwork compared to the conventional affine registration algorithms. As inaccurate affine pre-alignment of the image pair may impair the registration accuracy or impede the convergence of the deformable registration algorithm \cite{shen2019networks,zhou2014novel}, a comprehensive evaluation of the CNN-based affine registration methods should by no means be ignored. 

\subsection{Vision Transformer}
CNNs architecture generally has limitations in modelling explicit long-range dependencies due to the intrinsic inductive biases, \ie, weight sharing and locality, embedded into the architectural structure of CNNs. Recently, Dosovitskiy et al. \cite{dosovitskiy2020image} proposed a pioneering work, Vision Transformer (ViT), for image classification and proved that a pure transformer \cite{vaswani2017attention} architecture can attain a state-of-the-art performance. Compared to CNN-based approaches, ViT offers less image-specific inductive bias and has tremendous potential when training in large scale datasets. Wang et al. \cite{wang2021pyramid} develop a pyramid architectural design for a pure transformer model to imitate the multi-scale strategy in CNNs, achieving promising results in various computer vision tasks. Subsequent studies \cite{wang2021pvtv2,li2021localvit,guo2021cmt,dai2021coatnet,d2021convit,wu2021cvt,chu2021twins,chen2021visformer} further extend ViT to pyramid architectural design and introduce convolutions to ViT. These studies demonstrate that introducing moderate convolutional inductive bias to ViT improves the overall performance, especially for training with small datasets. Apart from pure ViT methods, Zhang et al. \cite{zhang2021learning} and Chen et al. \cite{chen2021vit} combine CNN encoder-decoder with transformer for deformable registration.

While CNNs have achieved remarkable success in deformable medical image registration, we argue that CNNs are not an ideal architecture for modelling and learning affine registration. In contrast to deformable image registration, affine registration is often used to mitigate and remove large linear misalignment, which is considered to be a global operation and contradicts the inductive bias embedded in the architectural structure of CNNs. Building on the insights of ViT and its variants \cite{dosovitskiy2020image,wang2021pyramid,wu2021cvt,d2021convit}, we depart from the CNNs architecture and propose a pure transformer-based method dedicated to 3D medical affine registration. 

\begin{figure*}[t]
\centering
	\begin{center}
        \includegraphics[width=1.0\linewidth]{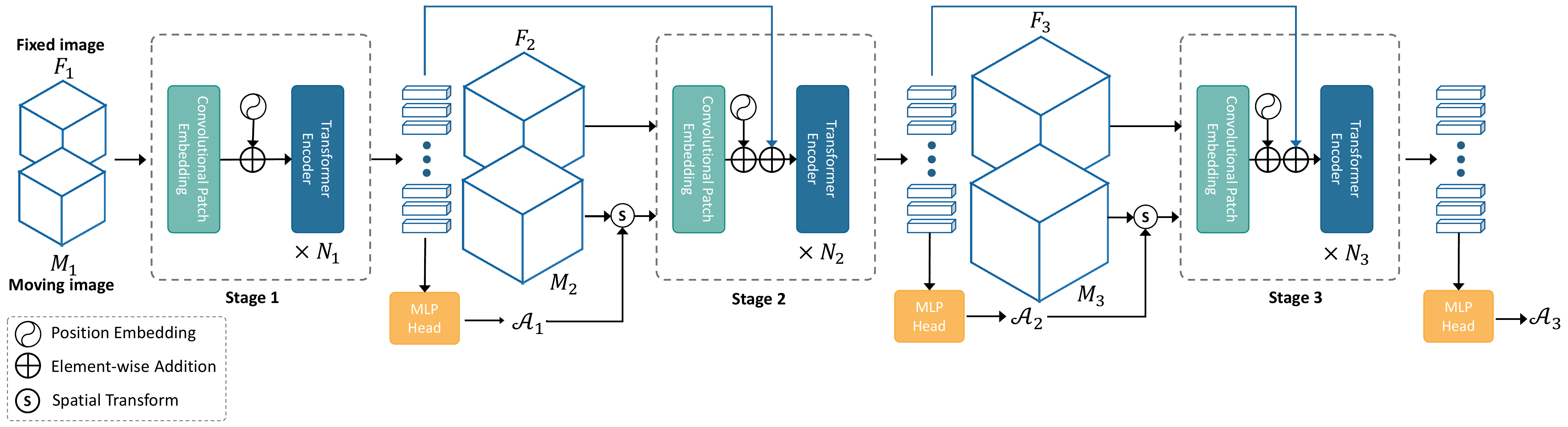}
	\end{center}
	\caption{Overview of the proposed Coarse-to-Fine Vision Transformer (C2FViT). The entire model is divided into three stages, solving the affine registration in a coarse-to-fine manner.}
	\label{fig:overview}
\end{figure*}

\section{Method}
Let $F$, $M$ be fixed and moving volumes defined over a $n$-D mutual spatial domain $\Omega \subseteq \mathbb{R}^n$. In this paper, we focus on 3D affine medical image registration, \ie, $n = 3$ and $\Omega \subseteq \mathbb{R}^3$. For simplicity, we further assume that $F$ and $M$ are single-channel, grayscale images. Our goal is to learn the optimal affine matrix that align $F$ and $M$. Specifically, we parametrized the affine registration problem as a function $f_\theta(F, M) = \mathcal{A}$ using a coarse-to-fine vision transformer (C2FViT), where $\theta$ is a set of learning parameters and $\mathcal{A}$ represents the predicted affine transformation matrix. 

\subsection{Coarse-to-fine Vision Transformer (C2FViT)}
The overall pipeline of our method is depicted in figure \ref{fig:overview}. Our method has been divided into $L$ stages that solves the affine registration in a coarse-to-fine manner with an image pyramid. All stages share an identical architecture consisting of a \emph{convolutional patch embedding} layer and $N_i$ transformer encoder blocks, where $N_i$ denotes the number of transformer blocks in stage $i$. Each transformer encoder block consists of an alternating multi-head self-attention module and a \emph{convolutional feed-forward layer}, as depicted in figure \ref{fig:compare}. We use $L = 3$ and $N_i = 4$ for each stage $i$ throughout this paper. Specifically, we first create the input pyramid by downsampling the input $F$ and $M$ with trilinear interpolation to obtain $F_i \in \{ F_1, F_2, \ldots ,F_L \}$ (and $M_i \in \{ M_1, M_2, \ldots ,M_L \}$), where $F_i$ represents the downsampled $F$ with a scale factor of $0.5^{L-i}$ and $F_L = F$. We then concatenate $F_i$ and $M_i$, and the concatenated input is subjected to the convolutional patch embedding layer. Different from the prior Transformer-based architectures \cite{dosovitskiy2020image,wang2021pyramid,wu2021cvt,d2021convit}, we prune all the layer normalization operations as we did not observe noticeable effects on the image registration performance in our experiments. Next, a stack of $N_i$ transformer encoder blocks take as input the image patch embedding map and output the feature embedding of the input. C2FViT solves the affine registration problem in a coarse-to-fine manner, and the intermediate input moving image $M_i$ is transformed via \emph{progressive spatial transformation}. Additionally, for stage $i > 1$, a residual connection from the output embeddings (tokens) of the previous stage $i-1$ is added to the patch embeddings of the current stage $i$. Finally, the estimated affine matrix $\mathcal{A}_L$ of the final stage is adopted as the output of our model $f_\theta$.

\vspace{-12 pt}
\subsubsection{Locality of C2FViT}
While the ViT model \cite{dosovitskiy2020image} excels in modelling long-range dependencies within a sequence of non-overlapping image patches due to the self-attention mechanism, the vision transformer model lacks locality mechanisms to model the relationship between the input patch and its neighbours. Therefore, we follow \cite{li2021localvit,wu2021cvt,wang2021pvtv2} to add locality to our transformers in C2FViT. Specifically, we mainly improve the transformer in two aspects: patch embedding and feed-forward layer. 

As shown in figure \ref{fig:overview}, we depart from the linear patch embedding approach \cite{dosovitskiy2020image} and adopt convolutional patch embedding \cite{wu2021cvt,wang2021pvtv2} instead. The goal of the convolutional patch embedding layer is to convert the input images into a sequence of overlapping patch embeddings. Formally, given a concatenated input $I \in \mathbb{R}^{H \times W \times D \times C}$, where $H$, $W$ and $D$ denote the spatial dimension of $I$, and $C$ is the number of channels, the convolutional patch embedding layer utilizes a 3D convolution layer to compute the patch embedding map $\mathbf{Z} \in \mathbb{R}^{H_i \times W_i \times D_i \times d}$ of $I$. Specifically, the kernel size, stride, number of zero-paddings and number of feature maps of the 3D convolution layer are denoted as $k^3$, $s$, $p$ and $d$, respectively. Next, the patch embedding map $\mathbf{Z}$ is then flattened into a sequence of patch embeddings (tokens) $\{\hat{\mathbf{Z}}_i \in \mathbb{R}^d |i=1, \ldots, N \}$, where $N = H_i W_i D_i$ and $d$ is the embedding dimension. The patch embeddings can be aggregated into a matrix $\hat{\mathbf{Z}} \in \mathbb{R}^{N \times d}$. We restrict the number of patches $N$ to $4096$ and the embedding dimension $d$ to $256$ for all convolutional patch embedding layers in C2FViT by varying the stride $s$ of the convolution layer, \ie, $s = (\frac{H}{16},\frac{W}{16},\frac{D}{16})$. Moreover, we enforce the window overlapping to the sliding window of the convolution operation by setting $k$ to $2s-1$, and pad the feature with zeros ($p= \lfloor \frac{k}{2} \rfloor$). In contrast to the linear patch embedding in ViT, the convolutional patch embedding in C2FViT helps model local spatial context and features across the fixed and moving images. It also provides flexibility to adjust the number and feature dimensions of patch embeddings. On the other hand, the feed-forward layer in ViT consists of a MLP block with two hidden layers. In the transformer encoder, the feed-forward layer is the only local and translation equivariance. Since the feed-forward layer in ViT is applied to the patch embeddings map in a patch-wise manner, it lacks a local mechanism to model the relationship between adjacent patch embeddings. As such, we add a $3 \times 3 \times 3$ depth-wise convolution layer in between two hidden layers of a MLP block in the feed-forward layer of C2FViT \cite{wang2021pvtv2,li2021localvit}. The depth-wise convolution further introduces locality into the transformer encoder of C2FViT.

\vspace{-12pt}
\subsubsection{Global Connectivity of C2FViT}
Transformers excel in modelling long-range dependencies within a sequence of embedding owing to their self-attention mechanism. In contrast to existing CNN-based affine registration approaches, the misalignment and the global relationship between the fixed and moving images can be captured and modelled by the similarity between the projected query-key pairs in transformer encoders of C2FViT, yielding the attention score for each patch embedding. Specifically, the query $\mathbf{Q}$, key $\mathbf{K}$, and value $\mathbf{V}$ are a linearly projection of the patch embeddings (tokens), \ie,  $\mathbf{Q}=\hat{\mathbf{Z}}\mathbf{W}^Q$, $\mathbf{K}=\hat{\mathbf{Z}}\mathbf{W}^K$ and $\mathbf{V}=\hat{\mathbf{Z}}\mathbf{W}^V$. We further extend the self-attention module to a multi-head self-attention (MHA) module \cite{vaswani2017attention}. Given the number of attention heads is $h$, the linear projection matrices $\mathbf{W}^Q_j$, $\mathbf{W}^K_j$ and $\mathbf{W}^V_j$ for each attention head $j$ are the same size, \ie,  $\mathbf{W}^Q_j$, $\mathbf{W}^K_j$, $\mathbf{W}^V_j \in \mathbb{R}^{d \times d_h}$ and $d_h = \frac{d}{h}$. Following the self-attention mechanism \cite{vaswani2017attention,dosovitskiy2020image} in the original transformer, our attention operation for attention head $j$ is computed as:
\begin{equation}\label{eq:attention}
	{\rm Attention}(\mathbf{Q}_j, \mathbf{K}_j, \mathbf{V}_j) = {\rm Softmax}(\frac{\mathbf{Q}_j \mathbf{K}_j^\mathsf{T}}{\sqrt{d_h}})\mathbf{V}_j
\end{equation}
\noindent where $d_h$ is the embedding dimension for the attention head. At the end, the attended embeddings of all attention heads are concatenated and linear projected by a matrix $\mathbf{W}^O \in \mathbb{R}^{d \times d}$. In this study, we employ $h=2$ attention heads and $d=256$ embedding dimension for all the transformer encoders.

\subsubsection{Progressive Spatial Transformation}
We adopt the multiresolution strategy into our architectural design. Specifically, a classification head, which is implemented by two successive multilayer perceptrons (MLP) layers with the hyperbolic tangent ({\rm Tanh}) activation function, is appended at the end of each stage in C2FViT. The classification head takes as input the averaged patch-wise patch embedding and outputs a set of affine transformation parameters. In the intermediate stage $i$, the derived affine matrix is used to progressively transform the moving image $M_{i+1}$ with a spatial transformer \cite{jaderberg2015spatial}. The warped moving image $M_{i+1}$ is then concatenated with fixed image $F_{i+1}$ and taken as input for stage $i+1$. With the proposed progressive spatial transformation, the linear misalignment of the input images can easily be eliminated with low-resolution input, and the transformers from the higher level can focus on the complex misalignment between the input image pair, reducing the complexity of the problem at the higher stages.

\subsection{Decoupled Affine Transformation}

While directly estimating the affine matrix is feasible \cite{shao2021weakly,zhao2019unsupervised,hu2018label}, this transformation model cannot generalize to other parametric registration methods as the affine matrix cannot decompose into a set of linear geometric transformation matrices, \ie, translation, rotation, scaling and shearing. In the transformation model of C2FViT, we take a step further and utilize C2FViT to predict a set of geometric transformation parameters instead of directly estimating the affine matrix. Formally, the affine registration problem is reduced to $f_\theta(F, M) = [\bm{t}, \bm{r}, \bm{s}, \bm{h}]$, where $\bm{t}, \bm{r}, \bm{s}, \bm{h} \in \mathbb{R}^3$ represent the translation, rotation, scaling and shearing parameters. Given $\mathcal{T}$, $\mathcal{R}$, $\mathcal{S}$ and $\mathcal{H}$, the resulting affine matrix $\mathcal{A}$ can be derived by a set of geometric transformation matrices via matrix multiplication as $\mathcal{A} = \mathcal{T} \cdot \mathcal{R} \cdot \mathcal{S} \cdot \mathcal{H}$, where $\mathcal{T}$, $\mathcal{R}$, $\mathcal{S}$ and $\mathcal{H}$ denote the translation, rotation, scaling and shearing transformation matrices derived by the corresponding geometric transformation parameters ($\bm{t}$, $\bm{r}$, $\bm{s}$ and $\bm{h}$), respectively. Our proposed transformation model can easily be transferred to other parametric registration settings by pruning or modifying undesired geometric transformation matrices. For instance, our C2FViT can be applied to rigid registration by removing the scaling and shearing matrices. Furthermore, our transformation model is capable of geometrical constraints, reducing the searching space of the model during optimization. In this work, the output geometric transformation parameters are constrained as follows: rotation and shearing parameters are constrained between $-\pi$ and $+\pi$, the translation parameters are constrained between -50\% and +50\% of the maximum spatial resolution, and the scaling parameters are constrained between 0.5 and 1.5. In this paper, we use the center of mass of the input instead of the geometric center for rotation and shearing. The center of mass $c_{I}$ of the image $I$ is defined as $c_{I} = \frac{\sum_{p \in \Omega} pI(p)}{\sum_{p \in \Omega} I(p)}$. If the background intensity of the image scan is non-zero, the origin of the rotation can be set to the geometric center of the image. 

\subsection{Unsupervised and Semi-supervised Learning}\label{sec:trans_model}
In contrast to the conventional affine registration methods, we parametrize the affine registration problem as a learning problem. Specifically, we formulate  the function $f_\theta(F, M) = \mathcal{A}_f$, where $f_\theta$ and $\mathcal{A}_f$ represent the C2FViT model and the output affine transformation matrix, respectively. Mathematically, our goal is to minimize the following equation:

\begin{equation}\label{eq:training}
	\theta^* = \argmin_\theta \Big[ \mathbb{E}_{(F,M) \in D} \; \mathcal{L} \big( F,M(\phi(\mathcal{A}_f) \big) \Big] ,
\end{equation}

\noindent where the $\theta$ is the learning parameters in C2FViT, fixed and moving images are randomly sampled from the training dataset $D$ and the loss function $\mathcal{L}$ measures the dissimilarity between the fixed image and the affine transformed moving image $M(\phi(\mathcal{A}_f))$. In our unsupervised learning setting, we use the negative NCC similarity measure with the similarity pyramid \cite{mok2020large} $\mathcal{L}_{sim}$ to quantify the distance between $F$ and $M(\phi(\mathcal{A}_f))$ such that $\mathcal{L}=\mathcal{L}_{sim}$ and $\mathcal{L}_{sim}$ is defined as:

\begin{equation}\label{eq:unsupervised}
	\mathcal{L}_{sim}(F,M(\phi)) = \sum_{i \in [1 .. L]} -\frac{1}{2^{(L-i)}} {\rm NCC}_w(F_i, M_i(\phi)),
\end{equation}

\noindent where $L$ denotes the number of image pyramid levels, ${\rm NCC}_w$ represents the local normalized cross-correlation with windows size $w^3$, and $(F_i, M_i)$ denotes the images in the image pyramid, \ie, $F_1$ is the image with the lowest resolution. In addition, our method is also capable of semi-supervised learning if the anatomical segmentation maps of the fixed and moving images are available in the training dataset. Given anatomical segmentation maps of fixed image $S_F$ and warped moving image $S_M(\phi)$, the semi-supervised C2FViT can be formulated by changing the similarity measure $\mathcal{L}$ in eq. \ref{eq:training} to $\mathcal{L}_{sim} + \lambda \mathcal{L}_{seg}$, where $\mathcal{L}_{seg}$ is defined as follows:

\begin{equation}\label{eq:semi_sup}
	\mathcal{L}_{seg}(S_F,S_M(\phi)) = \frac{1}{K} \sum_{i \in [1 .. K]} \Big( 1 - \frac{2(S_{F}^i \cap S_{M}^i(\phi))}{|S_{F}^i| + |S_{M}^i(\phi)|} \Big) 
\end{equation}

\noindent where $K$ denotes the number of anatomical structures. For the semi-supervised C2FViT, we utilize all available anatomical segmentations in our experiments. In this paper, we employ $L=3$ image pyramid levels and $\lambda=0.5$.

\begin{figure}[t]
	\begin{center}
        \includegraphics[width=1.0\linewidth]{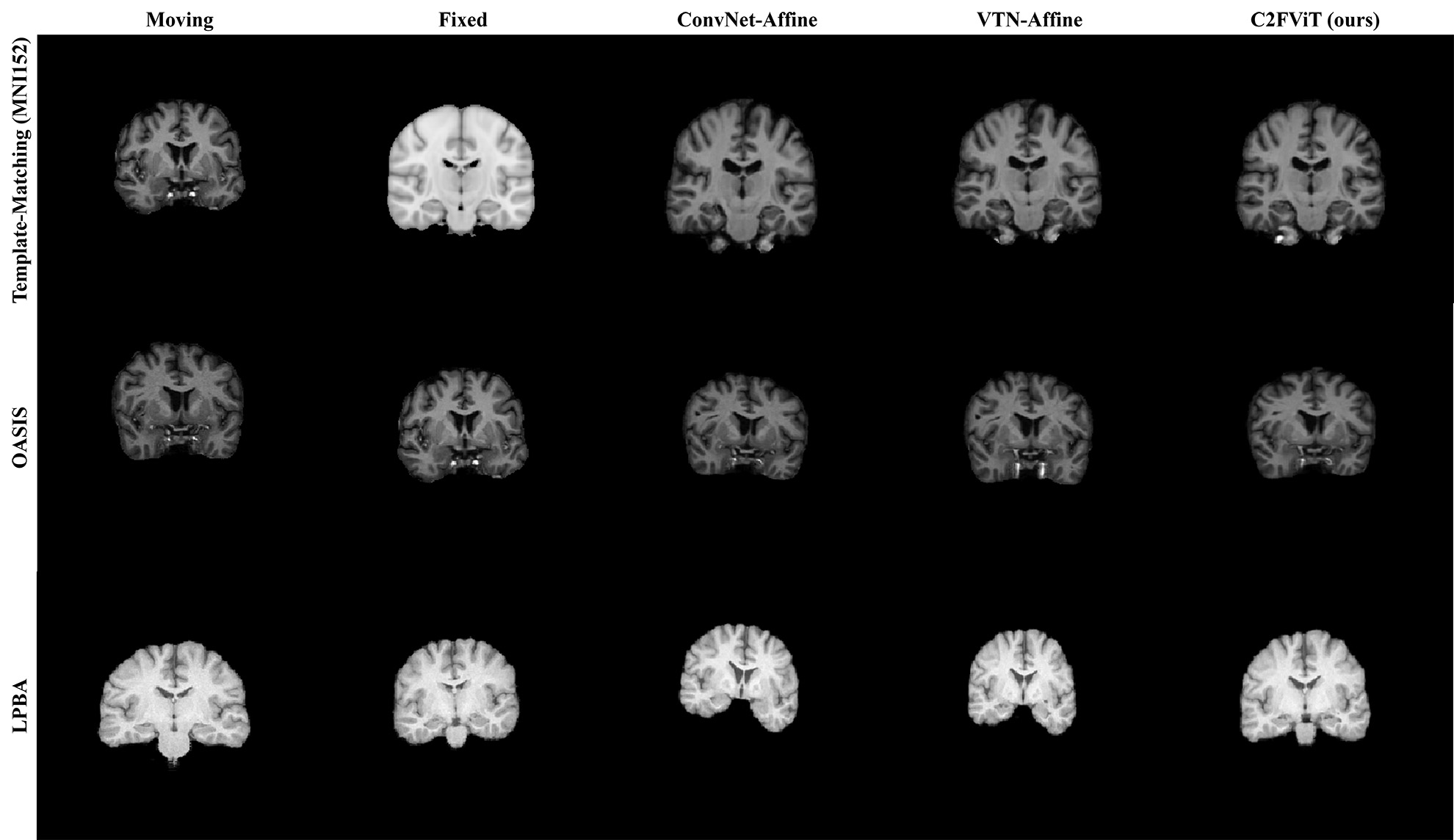}
	\end{center}
	\caption{Example coronal MR slices from the atlases (fixed images), moving images, resulting warped images for ConvNet-Affine, VTN-Affie and our method without center of mass initialization.}
	\label{fig:example_result}
\end{figure}

\section{Experiments}
\subsection{Data and Pre-processing}

We evaluated our method on brain template-matching normalization and atlas-based registration using 414 T1-weighted brain MRI scans from the OASIS dataset \cite{marcus2007open} and 40 brain MRI scans from the LPBA dataset \cite{shattuck2008construction}. For the OASIS dataset, we resampled and padded all MRI scans to $256 \times 256 \times 256$ with the same resolution ($1mm \times 1mm \times 1mm$) followed by standard preprocessing steps, including motion correction, skull stripping and subcortical structure segmentation, for each MRI scan using FreeSurfer \cite{fischl2012freesurfer}. For the LPBA dataset, the MRI scans are skull-stripped, and the manual delineation of the subcortical structures are provided. All brain MRI scans in our experiments are in native space, except the MNI152 brain template. We split the OASIS dataset into 255, 10 and 149 volumes for training, validation, and test sets, respectively. For the LPBA dataset, we included all 40 scans as the test set. 

\begin{table*}[t]
	\centering
	\resizebox{\textwidth}{!}{%
		\begin{tabular}{ccccccccccccccc}
			\toprule[1.5pt]
			\multirow{2}{*}{Method} & \multirow{2}{*}{\#Param} & \multicolumn{4}{c}{Template-Matching Normalization (MNI152)} & \multicolumn{4}{c}{Atlas-Based Registration (OASIS)} & \multicolumn{4}{c}{Atlas-Based Registration (OASIS$_{train}$ $\Rightarrow$ LPBA$_{test}$)}\\
			\cmidrule(lr){3-6}\cmidrule(lr){7-10}\cmidrule(lr){11-14}
			 & & \rule{1pt}{0ex} DSC$_4$ $\uparrow$ & DSC30$_4$ $\uparrow$ & HD95$_4$ $\downarrow$ & $\textnormal{T}_{test}$ $\downarrow$ & \rule{1pt}{0ex} DSC$_{23}$ $\uparrow$ & DSC30$_{23}$ $\uparrow$ & HD95$_{23}$ $\downarrow$ & $\textnormal{T}_{test}$ $\downarrow$ & \rule{1pt}{0ex} DSC$_3$ $\uparrow$ & DSC30$_3$ $\uparrow$ & HD95$_3$ $\downarrow$ & $\textnormal{T}_{test}$ $\downarrow$ \\
			\midrule[1pt]
			Initial \hspace{0.1cm} & - & 0.14 $\pm$ 0.12 & 0.02 $\pm$ 0.02 & 29.26 $\pm$ 11.33 & - & 0.18 $\pm$ 0.14 & 0.06 $\pm$ 0.02 & 15.53 $\pm$ 6.77 & - & 0.33 $\pm$ 0.06 & 0.26 $\pm$ 0.03 & 12.43 $\pm$ 4.65 & - \\
			\midrule
            ConvNet-Affine \cite{de2019deep} \hspace{0.1cm} & 14.7 M & 0.65 $\pm$ 0.08 & 0.56 $\pm$ 0.06 & 6.14 $\pm$ 1.33 & 0.12 $\pm$ 0.09 s & 0.57 $\pm$ 0.07 & 0.48 $\pm$ 0.05 & 4.10 $\pm$ 1.01 & 0.09 $\pm$ 0.06 s & 0.36 $\pm$ 0.07 & 0.28 $\pm$ 0.03 & 11.58 $\pm$ 4.99 & 0.11 $\pm$ 0.08 s \\
            VTN-Affine \cite{zhao2019unsupervised} \hspace{0.1cm} & 14.0 M & 0.67 $\pm$ 0.06 & 0.60 $\pm$ 0.05 & 5.80 $\pm$ 1.01 & \textbf{2e-3} $\pm$ 4e-4 s & 0.57 $\pm$ 0.08 & 0.48 $\pm$ 0.06 & 4.18 $\pm$ 1.08 & \textbf{3e-3} $\pm$ 8e-4 s & 0.31 $\pm$ 0.06 & 0.24 $\pm$ 0.03 & 14.99 $\pm$ 5.34 & \textbf{2e-3} $\pm$ 6e-4 s \\
            C2FViT (ours) \hspace{0.1cm} & 15.2 M & \textbf{0.71} $\pm$ 0.06 & \textbf{0.64} $\pm$ 0.04 & \textbf{5.17} $\pm$ 0.81 & 0.09 $\pm$ 0.03 s & \textbf{0.64} $\pm$ 0.06 & \textbf{0.57} $\pm$ 0.05 & \textbf{3.33} $\pm$ 0.77 & 0.08 $\pm$ 0.01 s & \textbf{0.47} $\pm$ 0.04 & \textbf{0.42} $\pm$ 0.02 & \textbf{6.55} $\pm$ 1.60 & 0.14 $\pm$ 0.06 s \\

			\bottomrule[1.5pt]
		\end{tabular}
	}
	\caption{Quantitative results of template-matching normalization and atlas-based registration \emph{without center of mass initialization}. The subscript of each metric indicates the number of anatomical structures involved. $\uparrow$: higher is better, and $\downarrow$: lower is better. Initial: initial results in native space without registration.}
	\label{tab:main_result}
\end{table*}

\begin{table*}[t]
	\centering
	\resizebox{\textwidth}{!}{%
		\begin{tabular}{ccccccccccccccc}
			\toprule[1.5pt]
			\multirow{2}{*}{Method} & \multirow{2}{*}{\#Param} & \multicolumn{4}{c}{Template-Matching Normalization (MNI152)} & \multicolumn{4}{c}{Atlas-Based Registration (OASIS)} & \multicolumn{4}{c}{Atlas-Based Registration (OASIS$_{train}$ $\Rightarrow$ LPBA$_{test}$)}\\
			\cmidrule(lr){3-6}\cmidrule(lr){7-10}\cmidrule(lr){11-14}
			 & & \rule{1pt}{0ex} DSC$_4$ $\uparrow$ & DSC30$_4$ $\uparrow$ & HD95$_4$ $\downarrow$ & $\textnormal{T}_{test}$ $\downarrow$ & \rule{1pt}{0ex} DSC$_{23}$ $\uparrow$ & DSC30$_{23}$ $\uparrow$ & HD95$_{23}$ $\downarrow$ & $\textnormal{T}_{test}$ $\downarrow$ & \rule{1pt}{0ex} DSC$_3$ $\uparrow$ & DSC30$_3$ $\uparrow$ & HD95$_3$ $\downarrow$ & $\textnormal{T}_{test}$ $\downarrow$ \\
			\midrule[1pt]
			Initial (CoM) \hspace{0.1cm} & - & 0.49 $\pm$ 0.11 & 0.35 $\pm$ 0.06 & 11.03 $\pm$ 3.48 & - & 0.45 $\pm$ 0.12 & 0.29 $\pm$ 0.06 & 6.97 $\pm$ 2.89 & - & 0.45 $\pm$ 0.04 & 0.41 $\pm$ 0.01 & 6.87 $\pm$ 1.69 & - \\
			\midrule
			Elastix \cite{klein2009elastix} \hspace{0.1cm} & - & 0.73 $\pm$ 0.07 & 0.64 $\pm$ 0.06 & 5.01 $\pm$ 1.44 & 6.6 $\pm$ 0.2 s & 0.63 $\pm$ 0.09 & 0.52 $\pm$ 0.08 & 3.89 $\pm$ 1.72 & 6.3 $\pm$ 0.2 s & \textbf{0.55} $\pm$ 0.02 & \textbf{0.53} $\pm$ 0.02 & 4.11 $\pm$ 1.01 & 6.4 $\pm$ 0.2 s \\
			ANTs \cite{avants2009advanced} \hspace{0.1cm} & - & 0.74 $\pm$ 0.06 & 0.67 $\pm$ 0.05 & 4.65 $\pm$ 0.57 & 38.2 $\pm$ 3.2 s & 0.67 $\pm$ 0.08 & 0.58 $\pm$ 0.08 & 3.27 $\pm$ 1.56 & 37.7 $\pm$ 2.5 s & 0.54 $\pm$ 0.03 & 0.50 $\pm$ 0.02 & 4.53 $\pm$ 1.38 & 46.6 $\pm$ 15.3 s \\
            \midrule
            ConvNet-Affine \cite{de2019deep} \hspace{0.1cm} & 14.7 M & 0.70 $\pm$ 0.06 & 0.63 $\pm$ 0.05 & 5.28 $\pm$ 0.68 & 0.12 $\pm$ 0.08 s & 0.62 $\pm$ 0.06 & 0.55 $\pm$ 0.05 & 3.43 $\pm$ 0.91 & 0.10 $\pm$ 0.07 s & 0.45 $\pm$ 0.04 & 0.41 $\pm$ 0.01 & 7.46 $\pm$ 1.87 & 0.11 $\pm$ 0.08 s \\
            VTN-Affine \cite{zhao2019unsupervised} \hspace{0.1cm} & 14.0 M & 0.71 $\pm$ 0.06 & 0.64 $\pm$ 0.05 & 5.11 $\pm$ 0.74 & 3e-3 $\pm$ 9e-4 s & 0.66 $\pm$ 0.06 & 0.59 $\pm$ 0.06 & 3.02 $\pm$ 0.81 & \textbf{2e-3} $\pm$ 7e-4 s & 0.43 $\pm$ 0.04 & 0.39 $\pm$ 0.02 & 8.02 $\pm$ 2.23 & \textbf{2e-3} $\pm$ 6e-4 s \\
            C2FViT (ours) \hspace{0.1cm} & 15.2 M & 0.72 $\pm$ 0.06 & 0.65 $\pm$ 0.05 & 4.99 $\pm$ 0.75 & 0.12 $\pm$ 0.04 s &  0.66 $\pm$ 0.05 & 0.61 $\pm$ 0.04 & 2.96 $\pm$ 0.54 & 0.09 $\pm$ 0.02 s & 0.54 $\pm$ 0.03 & 0.51 $\pm$ 0.04 & \textbf{4.06} $\pm$ 1.12 & 0.12 $\pm$ 0.04 s \\
            \midrule
            ConvNet-Affine-semi \cite{de2019deep} \hspace{0.1cm} & 14.7 M & 0.73 $\pm$ 0.06 & 0.66 $\pm$ 0.04 & 4.94 $\pm$ 0.76 & 0.12 $\pm$ 0.09 s & 0.63 $\pm$ 0.06 & 0.56 $\pm$ 0.06 & 3.46 $\pm$ 0.96 & 0.10 $\pm$ 0.07s & 0.43 $\pm$ 0.03 & 0.40 $\pm$ 0.02 & 6.90 $\pm$ 1.52 & 0.12 $\pm$ 0.08 s \\
            VTN-Affine-semi \cite{zhao2019unsupervised} \hspace{0.1cm} & 14.0 M & 0.75 $\pm$ 0.05 & \textbf{0.70} $\pm$ 0.04 & 4.65 $\pm$ 0.66 & \textbf{2e-3} $\pm$ 6e-4 s & 0.68 $\pm$ 0.05 & 0.62 $\pm$ 0.04 & 2.94 $\pm$ 0.64 & \textbf{2e-3} $\pm$ 8e-4 s & 0.44 $\pm$ 0.04 & 0.40 $\pm$ 0.02 & 7.27 $\pm$ 1.96 & \textbf{2e-3} $\pm$ 1e-3 s \\
            C2FViT-semi (ours) \hspace{0.1cm} & 15.2 M & \textbf{0.76} $\pm$ 0.05 & \textbf{0.70} $\pm$ 0.04 & \textbf{4.60} $\pm$ 0.69 & 0.13 $\pm$ 0.05 s & \textbf{0.69} $\pm$ 0.04 & \textbf{0.64} $\pm$ 0.04 & \textbf{2.81} $\pm$ 0.55 & 0.08 $\pm$ 0.02 s & 0.51 $\pm$ 0.03 & 0.47 $\pm$ 0.04 & 4.58 $\pm$ 1.71 & 0.13 $\pm$ 0.05 s \\

			\bottomrule[1.5pt]
		\end{tabular}
	}
	\caption{Quantitative results on template-matching normalization, OASIS and LPBA dataset \emph{with center of mass initialization}. The subscript of each metric indicates the number of anatomical structures involved. $\uparrow$: higher is better, and $\downarrow$: lower is better. Initial (CoM): initial results with the center of mass initialization. To our knowledge, ANTs and Elastix do not have a GPU implementation.}
	\label{tab:main_result_COM}
\end{table*}

We evaluated our method on two applications of brain registration: brain template-matching normalization to MNI152 space and atlas-based registration in native space. Brain template-matching normalization is a standard application in analyzing inter-subject images and a necessary pre-processing step in most deformable image registration methods. For the task of brain template-matching normalization, we affinely register all test scans in the OASIS dataset to an MNI152 (6$^\text{th}$ generation) brain template \cite{grabner2006symmetric,evans2012brain,fischl2012freesurfer}, which is derived from 152 structural images and averaged together after non-linear registration into the common MNI152 co-ordinate system. We train the learning-based methods with the training dataset of OASIS and the MNI152 template, which employ the MNI152 template as the fixed image and MRI scans from the training dataset as moving images. For the atlas-based registration task, we randomly select 3 and 2 scans from the test set of OASIS and LPBA datasets respectively as atlases. Then, we align the remaining MRI scans in the test set to the selected atlases within the same dataset. Note that in the atlas-based registration task, we train the learning-based methods with pairwise brain registration, which randomly samples two image scans as fixed and moving images, using only the training set of the OASIS dataset, \ie, the selected atlases and the MRI scans from the LPBA dataset were not involved in the training. 

Conventionally, affine registration methods often initialize the input images with center of mass (CoM) initialization by default \cite{mccormick2014itk}, which initializes the translation parameters using the CoM of the input images. Equivalently, the CoM initialization for learning-based methods can be achieved by translating the CoM of the moving image to the CoM of the fixed image. We evaluated our method with and without the CoM initialization, and the results are listed in table \ref{tab:main_result} and table \ref{tab:main_result_COM}, respectively. 

\subsection{Measurement}
\begin{figure*}[t]
	\begin{center}
        \includegraphics[width=1.0\linewidth]{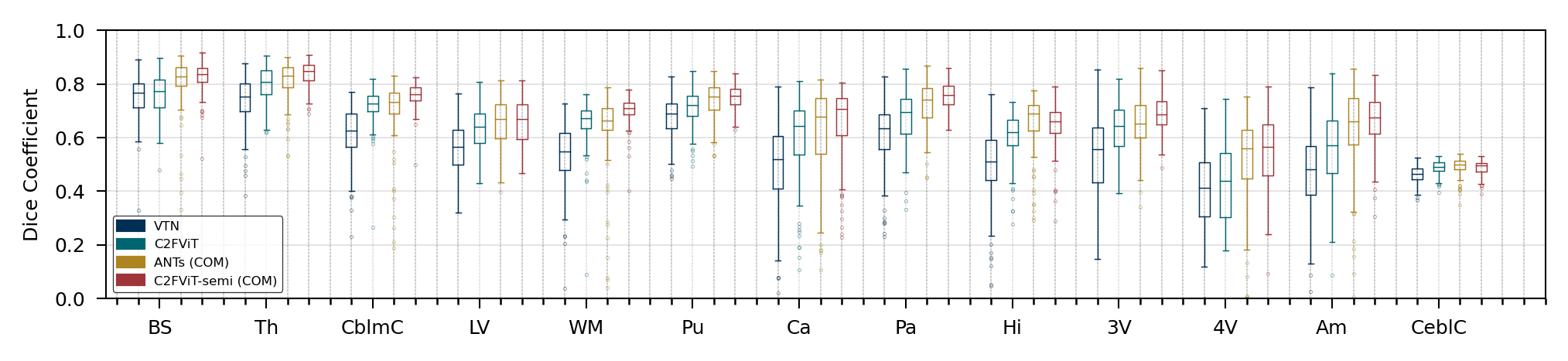}
	\end{center}
	\caption{Boxplots illustrating Dice scores of each anatomical structure for C2FViT, VTN and ANTs in the atlas-based registration with the OASIS dataset. The left and right hemispheres of the brain are combined into one structure for visualization. The brain stem (BS), thalamus (Th), cerebellum cortex (CblmC), lateral ventricle (LV), cerebellum white matter (WM), putamen (Pu), caudate (Ca), pallidum (Pa), hippocampus (Hi), 3rd ventricle (3V), 4th ventricle (4V), amygdala (Am), and cerebral cortex (CeblC) are included. Methods with (CoM) postfix are trained and tested on MRI scans with the center of mass initialization. }
	\label{fig:box_plot}
\end{figure*}

To quantify the registration performance of an affine registration algorithm, we register each subject to an atlas or MNI152 template, propagate the subcortical structure segmentation map using the resulting affine transformation matrix, and measure the volume overlap using the Dice similarity coefficient (DSC) and 30\% lowest DSC of all cases (DSC30). We also measure the 95\% percentile of the Hausdorff distance (HD95) of the segmentation map to represent the reliability of the registration algorithm. In the brain template-matching normalization task, 4 subcortical structures, \ie, caudate, cerebellum, putamen and thalamus, are included in the evaluation. In the atlas-based registration with the OASIS dataset, 23 subcortical structures are included, as shown in the boxplot in figure \ref{fig:box_plot}. For the atlas-based registration with the LPBA dataset, we utilize all manual segmentation of the brain scan, including cerebrospinal fluid (CSF), gray matter (GM) and white matter (WM), for evaluation.

\subsection{Baseline Methods}
We compare our method with two state-of-the-art conventional affine registration methods (ANTs\cite{avants2009advanced} and Elastix \cite{klein2009elastix}) and two learning-based affine registration approaches (ConvNet-Affine \cite{de2019deep} and VTN-Affine \cite{zhao2019unsupervised}). Specifically, we use the ANTs affine registration implementation in the publicly available ANTs software package \cite{avants2011reproducible}, and we use the Elastix affine registration algorithm in the SimpleElastix toolbox \cite{marstal2016simpleelastix}. Both methods use a 3-level multi-resolution optimization strategy with adaptive gradient descent optimization and the mutual information as the similarity measure. For ConvNet-Affine and VTN-Affine, we follow their papers to implement their affine subnetworks. The initial number of feature channels for both methods is set to 16, and we follow the rules in their papers to define the growth of network depth and the hidden dimension of each convolution layer. By default, all learning-based methods are trained in an unsupervised manner with the similarity pyramid as described in eq. \ref{eq:unsupervised}. We also extend the unsupervised learning-based methods to semi-supervised variants using the same semi-supervised object function as our method, denoted as C2FViT-semi, ConvNet-Affine-semi and VTN-Affine-semi. 

\subsection{Implementation}
The learning-based methods, \ie, C2FViT, ConvNet-Affine and VTN-Affine, are developed and trained using Pytorch. All the methods are trained or executed on a standalone workstation equipped with an Nvidia TITAN RTX GPU and an Intel Core i7-7700 CPU. The learning-based approaches are trained with half-resolution image scans by downsampling the image scans with trilinear interpolation. Then, we apply the resulting affine transformation to the full-resolution image scans for evaluation. We adopt the Adam optimizer \cite{kingma2014adam} with a fixed learning rate of $1e^{-4}$ and batch size sets to 1 for all learning-based approaches. 

\subsection{Results}
\subsubsection{Registration accuracy and Robustness}

Table \ref{tab:main_result} shows the results of template-matching normalization and atlas-based registration of the learning-based methods \emph{without spatial initialization}. Figure \ref{fig:example_result} illustrates the qualitative results of all tasks without spatial initialization. The low initial Dice scores over all subjects, suggesting that there is a large misalignment within each test case. Our proposed method is significantly better than ConvNet-Affine and VTN-Affine in terms of DSC, DSC30 and HD95 over all three tasks, suggesting our method is robust and accurate in affine registration with large initial misalignment. We visualize the distribution of Dice scores for each subcortical structure as in the boxplot in figure \ref{fig:box_plot}. Compared to VTN-Affine, the C2FViT model achieves consistently better performance across all structures. 

Table \ref{tab:main_result_COM} shows the results of tasks with CoM initialization. This simple but effective initialization boosts the initial Dice scores from 0.14, 0.18 and 0.33 to 0.49, 0.45 and 0.45, respectively, implying that the initialization eliminates most of the misalignment due to translation. All three learning-based methods improve significantly on affine alignment with CoM initialization. For an unsupervised manner, our method achieves comparable Dice measures to the conventional methods (ANTs and Elastix), and slightly better than ConvNet-Affine and VTN-Affine. It is worth noting that VTN-Affine gains significant improvement in registration performance of template-matching and atlas-based registration (OASIS) under CoM initialization. Nevertheless, the validity of the initial registration should be questioned when the two images are acquired in different imaging modalities and hence, the registration performance without spatial initialization should be considered when evaluating the learning-based affine registration algorithm. With our proposed semi-supervised settings, our method C2FViT-semi achieves the best overall registration performance in the template-matching normalization and the atlas-based registration task on the OASIS dataset. 

\vspace{-12pt}
\subsubsection{Generalizability Analysis}
As shown in the results of the LPBA dataset in tables \ref{tab:main_result} and \ref{tab:main_result_COM}, ConvNet-Affine and VTN-Affine, using models trained on the OASIS dataset, fail spectacularly in the test set of LPBA, which obtain -5\% and -2\% loss in DSC with VTN-Affine, and +3\% and +0\% gain in DSC with ConvNet-Affine compared to initial results without registration and with spatial initialization, respectively. The results imply that their models cannot generalize well to an unseen dataset in practice regardless of spatial initialization. By contrast, our C2FViT model achieves a comparable registration performance to the conventional affine registration approaches ANTs and Elastix in the task with the LPBA dataset, reaching an average Dice score of 0.54 and HD95 of 4.06 in the task with the LPBA dataset, as shown in table \ref{tab:main_result_COM}. While the semi-supervised settings improve the dataset-specific performance of learning-based models in template-matching normalization and atlas-based registration with the OASIS dataset, the semi-supervised models are inferior to their unsupervised models in the LPBA dataset, indicating anatomical knowledge injected to the model with semi-supervision may not generalize well to unseen data beyond the training dataset.


\begin{table}[h]
	\begin{center}
		\scalebox{1.0}{
			\resizebox{0.45\textwidth}{!}{
					\begin{tabular}{l|lll|l}
						\toprule
						Methods              & DSC$_{23}$ & HD95$_{23}$ & $\textnormal{T}_{test}$ & \#Param  \\ 
						\hline
						Vanilla C2FViT-s1   & 0.61 & 3.53 & 0.05 $\pm$ 0.04 s  & 5.0 M    \\ 
						Vanilla C2FViT-s2   & 0.62 & 3.57 & 0.06 $\pm$ 0.05 s  & 10.0 M    \\ 
						\hline
						Vanilla C2FViT-s3   & 0.62 & 3.46 & 0.07 $\pm$ 0.02 s  & 15.2 M    \\ 
						
						\ \ \ \ \small{+Progressive Spatial Transformation}    & 0.64 \color{ForestGreen}\small \textbf{(+0.02)}   & 3.33 \color{ForestGreen}\small \textbf{(-0.13)}   & 0.08 $\pm$ 0.02 s  & 15.2 M     \\
						\ \ \ \ \small{+Center of Mass Initialization}    & 0.66 \color{ForestGreen}\small \textbf{(+0.02)}   & 2.96 \color{ForestGreen}\small \textbf{(-0.37)}   & 0.09 $\pm$ 0.02 s  & 15.2 M     \\
						\ \ \ \ \small{+Semi-supervision}    & 0.69 \color{ForestGreen}\small \textbf{(+0.03)}   & 2.81 \color{ForestGreen}\small \textbf{(-0.15)}   & 0.08 $\pm$ 0.02 s  & 15.2 M     \\


						\bottomrule
						\end{tabular}
					}}
	\end{center}
	\caption{Influence of the number of stages, progressive spatial transformation, center of mass initialization and the semi-supervised learning to the C2FViT model. The C2FViT with postfix -s$\{ n \}$ represents  the C2FViT model with an $n$-stage. }
	\label{tab:ablation}
\end{table}

\subsubsection{Runtime Analysis}
The average runtimes (denoted as $\textnormal{T}_{test}$) of all methods in the inference phase are reported in tables \ref{tab:main_result} and \ref{tab:main_result_COM}. We report the average registration time for each task. C2FViT, ConvNet-Affine and VTN-Affine are faster than the ANTs and Elastix by order of magnitude, thanks to the GPU acceleration and the effective learning formulation. Moreover, ANTs runtimes vary widely, as its convergence depends on the degree of initial misalignment of the task. On the other hand, Elastix runtimes are stable at around 6.6 seconds per alignment task because of the early stopping strategy used during the affine alignment. 

\begin{table}[h]
\centering
\footnotesize
\resizebox{0.45\textwidth}{!}{
    \begin{tabular}{l|cccc}
    \toprule
    & DSC$_{23}$ $\uparrow$ & DSC30$_{23}$ $\uparrow$ & HD95$_{23}$ $\downarrow$ & $\textnormal{T}_{test}$ $\downarrow$\\
    \hline
    C2FViT-direct  & 0.63 $\pm$ 0.06 & 0.55 $\pm$ 0.04 & 3.43 $\pm$ 0.73 & 0.02 $\pm$ 4e-3 s \\
    C2FViT-decouple & 0.64 $\pm$ 0.06 & 0.57 $\pm$ 0.05 & 3.33 $\pm$ 0.77 & 0.08 $\pm$ 0.01 s \\
    \bottomrule
    \end{tabular}
}
\caption{Influence of the proposed decoupled affine transmation model compared to the direct affine matrix estimation model.}
\label{tab:transformation_model_result}
\end{table}

\subsubsection{Ablation study}
Table \ref{tab:ablation} shows the ablation study results of C2FViT in the OASIS atlas-based registration task. The results suggest that the proposed progressive spatial transformation, CoM initialization and semi-supervised learning consistently improve the registration performance of C2FViT without adding extra learning parameters or significant computational burden to the model. Table \ref{tab:transformation_model_result} presents the results of C2FViT using two different transformation models in the OASIS atlas-based registration task. The proposed decoupled affine transformation model is slightly better than directly learning the affine matrix, in terms of registration performance, at the cost of registration runtime. Moreover, the decoupled affine transformation model can be easily adapted to other parametric registration methods by pruning or modifying the geometrical transformation matrices.

\section{Conclusion}
We have proposed a Coarse-to-Fine Vision Transformer dedicated to 3D affine medical image registration. Unlike prior works using CNN-based affine registration methods, our method leverages the global connectivity of the self-attention operator and moderates the locality of the convolutional feed-forward layer to encode the global orientations, spatial positions and long-term dependencies of the image pair to a set of geometric transformation parameters. Comprehensive experiments demonstrate that our method not only achieves superior registration performance over the existing CNN-based methods under data with large initial misalignment and is robust to an unseen dataset, but also our method with semi-supervision outperforms conventional methods in terms of dataset-specific and preserves the runtime advantage of learning-based methods. Nevertheless, there is still a gap between unsupervised learning-based approaches and conventional approaches. We believe that expanding the training dataset and introducing task-specific data augmentation techniques would likely lead to performance improvement. 

\newpage
{\small
\bibliographystyle{ieee_fullname}
\bibliography{egbib}
}
\newpage
\appendix



\section{Unsupervised and Semi-Supervised Learning}
\begin{figure}[h]
	\begin{center}
        \includegraphics[width=1.0\linewidth]{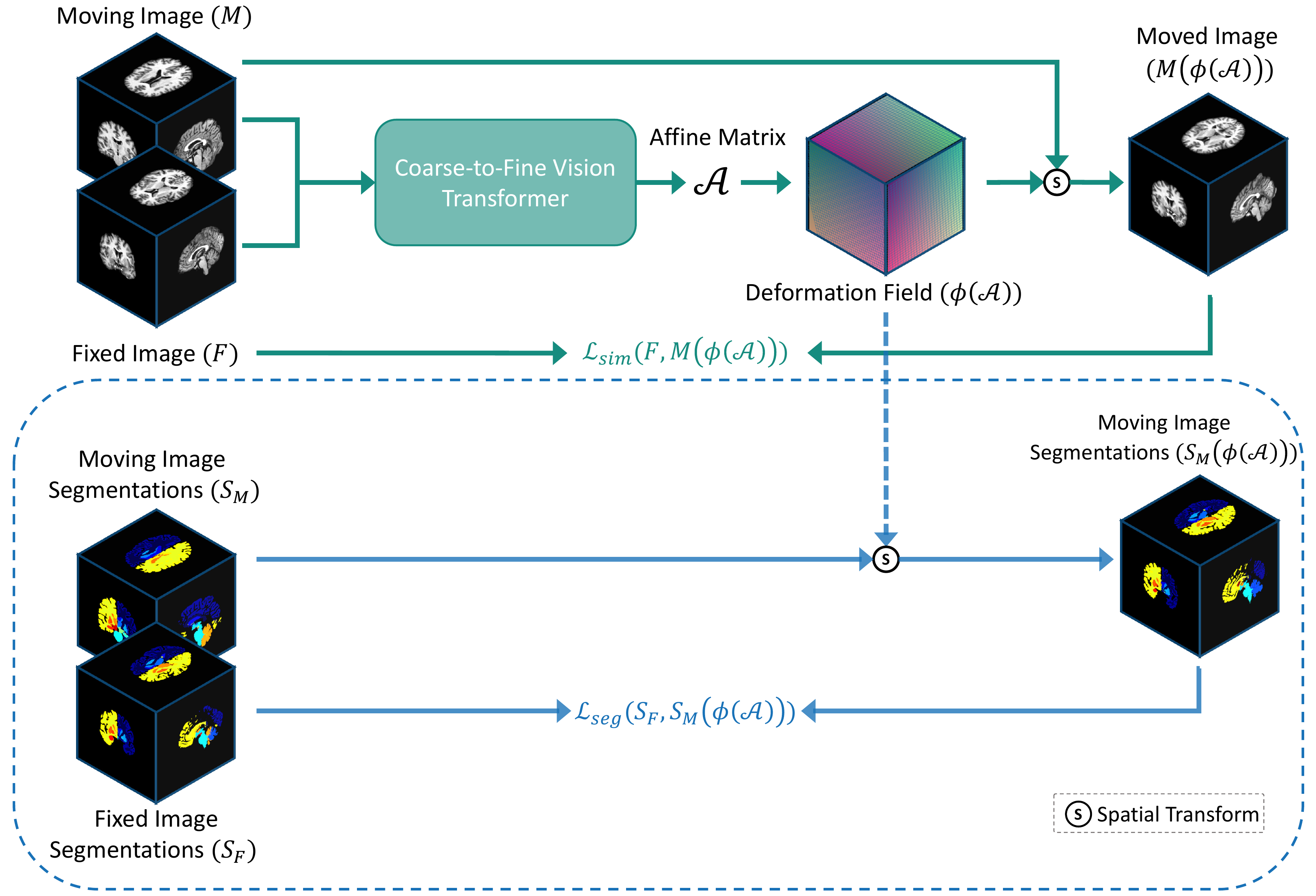}
	\end{center}
	\caption{Schematic representation of the unsupervised and semi-supervised learning scheme in the Coarse-to-Fine Vision Transformer. The unsupervised and semi-supervised learning schemes are highlighted in green and blue colours, respectively.}
	\label{fig:training}
\end{figure}

Figure \ref{fig:training} depicts the proposed unsupervised and semi-supervised training scheme of the Coarse-to-Fine Vision Transformer (C2FViT). The segmentation maps are only required in the training phrase under the semi-supervised training scheme.

\section{Affine Transformations}
The corresponding translation $\mathcal{T}$, rotation $\mathcal{R}$, scaling $\mathcal{S}$ and shearing $\mathcal{H}$ transformations derived by the geometric transformation parameters $t_x, t_y, t_z \in \bm{t}$, $r_x, r_y, r_z \in \bm{r}$, $s_x, s_y, s_z \in \bm{s}$ and $h_x, h_y, h_z \in \bm{h}$ are defined as follows:

\resizebox{0.4\textwidth}{!}{$\mathcal{T}=\begin{pmatrix} 
1 & 0 & 0 & t_x \\
0 & 1 & 0 & t_y \\
0 & 0 & 1 & t_z \\
0 & 0 & 0 & 1
\end{pmatrix},
\mathcal{R}_x=\begin{pmatrix} 
1 & 0 & 0 & 0 \\
0 & \cos(r_x) & \sin(r_x) & 0 \\
0 & -\sin(r_x) & \cos(r_x) & 0 \\
0 & 0 & 0 & 1
\end{pmatrix}$}

\resizebox{0.4\textwidth}{!}{$\mathcal{S}=\begin{pmatrix} 
s_{x} & 0 & 0 & 0 \\
0 & s_{y} & 0 & 0 \\
0 & 0 & s_{z} & 0 \\
0 & 0 & 0 & 1
\end{pmatrix},
\mathcal{R}_y=\begin{pmatrix} 
\cos(r_y) & 0 & -\sin(r_y) & 0 \\
0 & 1 & 0 & 0 \\
\sin(r_y) & 0 & \cos(r_y) & 0 \\
0 & 0 & 0 & 1
\end{pmatrix}$}

\resizebox{0.4\textwidth}{!}{$\mathcal{H}=\begin{pmatrix} 
1 & h_{xy} & h_{xz} & 0 \\
0 & 1 & h_{yz} & 0 \\
0 & 0 & 1 & 0 \\
0 & 0 & 0 & 1
\end{pmatrix},
\mathcal{R}_z=\begin{pmatrix} 
\cos(r_z) & -\sin(r_z) & 0 & 0 \\
\sin(r_z) & \cos(r_z) & 0 & 0 \\
0 & 0 & 1 & 0 \\
0 & 0 & 0 & 1
\end{pmatrix}$}
\newline

\noindent where rotation matrix $\mathcal{R}$ equals to  $\mathcal{R}_x \mathcal{R}_y \mathcal{R}_z$.


\section{Additional Implementation Details}
Table \ref{tab:config} summarizes the configurations of C2FViT at each stage. Specifically, the input resolution, stride in the convolutional patch embedding, number of transformer encoders, embedding size of each patch embedding, embedding size of the convolutional feed-forward layer and number of heads for the multi-head self-attention module are listed in the table.

\begin{table}[h]
	\centering
    \resizebox{0.4\textwidth}{!}{%
    \begin{tabular}{lcccccc}
    \toprule[1 pt]
    Stage & Input size & Stride & \# Encoders & Hidden size & MLP size & Heads \\
    \midrule[1 pt]
    Stage 1 & $32^3$ & $2^3$ & $4$ & $256$ & $512$ & $2$ \\
    Stage 2 & $64^3$ & $4^3$ & $4$ & $256$ & $512$ & $2$ \\
    Stage 3 & $128^3$ & $8^3$ & $4$ & $256$ & $512$ & $2$ \\
    \bottomrule[1 pt]
    \end{tabular}
    }
    \caption{Model configurations of Coarse-to-Fine Vision Transformer at each stage.}
    \label{tab:config}
\end{table}

\section{Additional Qualitative Results}
\begin{figure}[h]
	\begin{center}
        \includegraphics[width=1.0\linewidth]{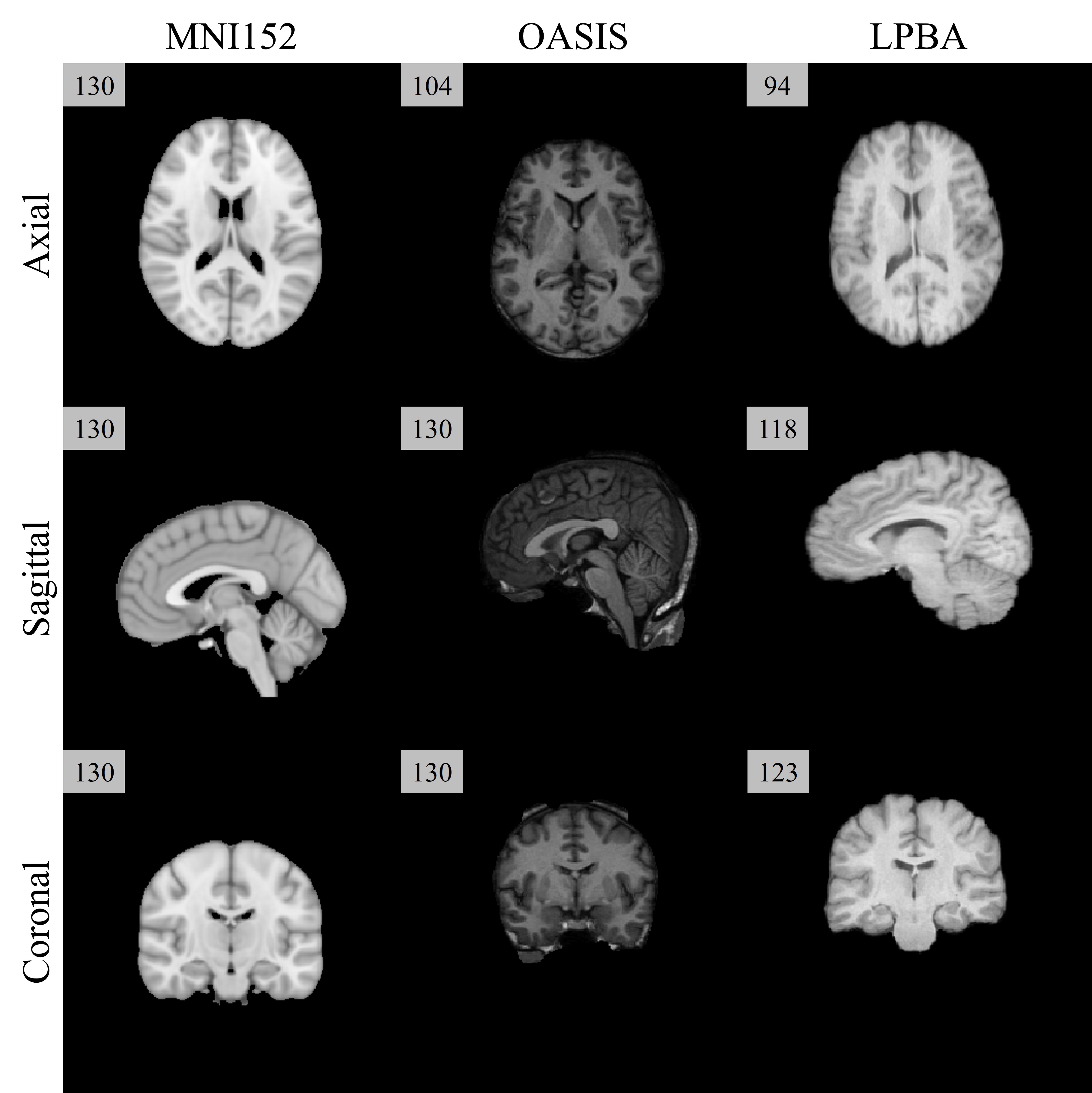}
	\end{center}
	\caption{Example axial, sagittal and coronal slices obtained from the MNI152 template, OASIS and LPBA brain MRI datasets. The corresponding slice number of each slice is highlighted at the top-left corner.}
	\label{fig:example_slice}
\end{figure}

\begin{figure*}[t]
	\begin{center}
        \includegraphics[width=1.0\linewidth]{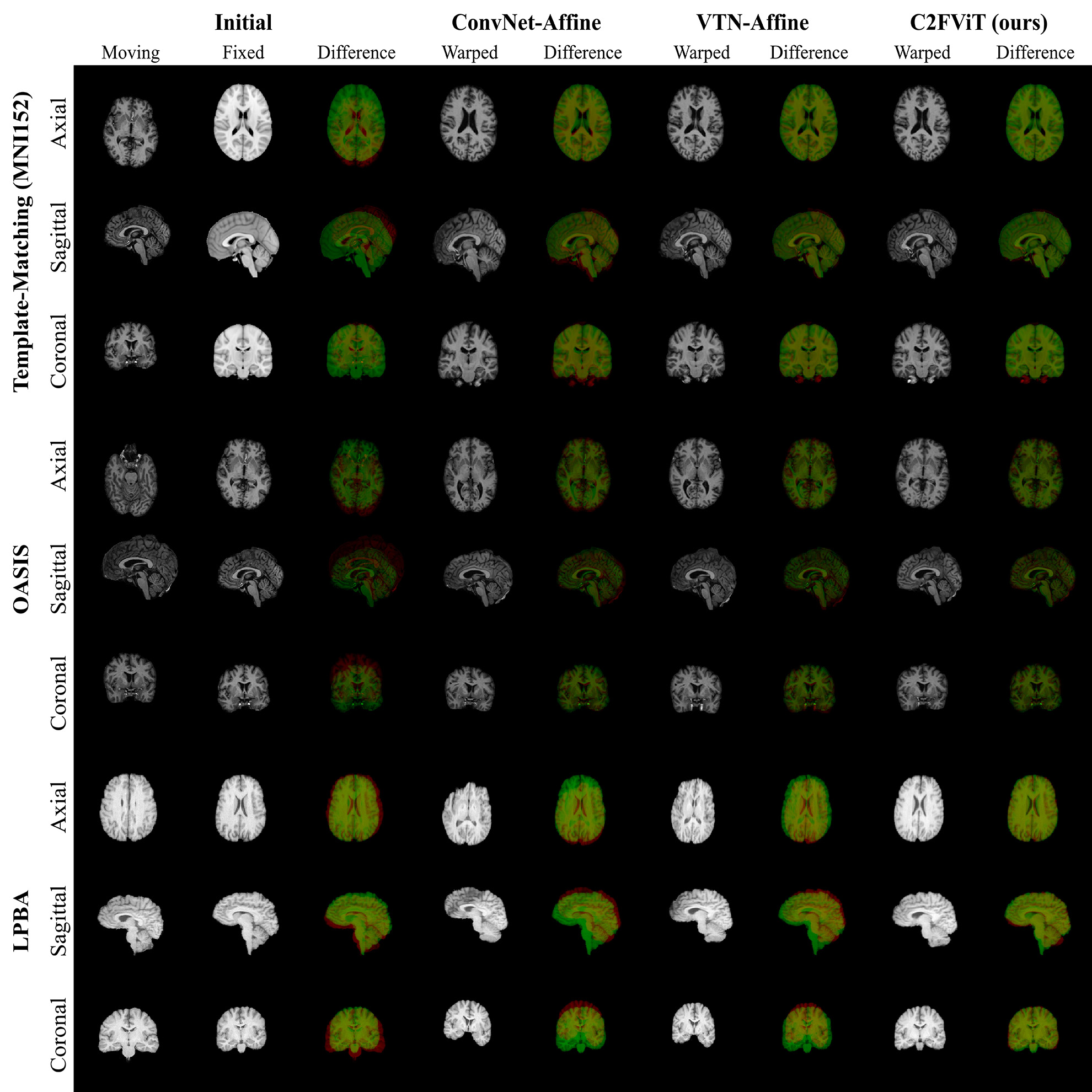}
	\end{center}
	\caption{Example axial, sagittal and coronal MR slices obtained from the moving images, atlases (fixed images), resulting warped images for ConvNet-Affine, VTN-Affie and our method without center of mass initialization. For better visualization, we depict a difference map for each method, in which the colour maps of fixed and warped moving images are set to black-green and black-red, respectively, and overlay the resulting warped moving image to fixed image.}
	\label{fig:detail_qual}
\end{figure*}

Figure \ref{fig:example_slice} shows example MR slices obtained from the MNI152 template, OASIS and LPBA datasets. As shown in the figure, there are significant spatial and structural differences across scans as all scans are in native space, except for the MNI152 template. The comprehensive qualitative results of template-matching normalization and atlas-based registration tasks with the OASIS and LPBA dataset of the learning-based methods \emph{without spatial initialization} are shown in figure \ref{fig:detail_qual}. 

\newpage
\section{Details of ANTs and Elastix}

\noindent The command and parameters we used for ANTs: \\
\begin{footnotesize}
\texttt{-d 3 -v 1 -t Affine[0.1]
\\ -m MI[<Fixed>,<Moving>,1,32,Regular,0.1] 
\\-c 200x200x200 
-f 4x2x1 
-s 2x1x0 
\\-o <OutFileSpec>}
\end{footnotesize}\\


\noindent The command and parameters we used for Elastix: \\
\begin{footnotesize}
\texttt{ef = sitk.ElastixImageFilter()
ef.SetFixedImage(sitk.ReadImage(<Fixed>))
ef.SetMovingImage(sitk.ReadImage(<Moving>))
pmap = sitk.GetDefaultParameterMap("affine")
ef.SetParameterMap(pmap)
\\ef.Execute()}
\end{footnotesize}

\end{document}